\documentclass{article}
\usepackage{ijcai25}

\usepackage{tikz}
\usepackage[edges]{forest}
\definecolor{hidden-draw}{RGB}{20,68,106}
\definecolor{hidden-pink}{RGB}{255,245,247}

\usepackage{times}
\usepackage{soul}
\usepackage{url}
\usepackage[hidelinks]{hyperref}
\usepackage[utf8]{inputenc}
\usepackage[small]{caption}
\usepackage{graphicx}
\usepackage{amsmath}
\usepackage{amsthm}
\usepackage{booktabs}
\usepackage{algorithm}
\usepackage{algorithmic}
\usepackage[switch]{lineno}

\usepackage{array}  % 使用 array 宏包来控制单元格对齐
\usepackage{enumitem}
\title{Neuro-Symbolic Artificial Intelligence: \\ Towards Improving the Reasoning Abilities of Large Language Models}

\author{
Xiao-Wen Yang$^{1,2*}$\and
Jie-Jing Shao$^{1*}$\and
Lan-Zhe Guo$^{1,3*}$\and
Bo-Wen Zhang$^{1,3}$\and \\
Zhi Zhou$^{1}$\and 
Lin-Han Jia$^{1}$\and 
Wang-Zhou Dai$^{1,3}$\And 
Yu-Feng Li$^{1,2\textsuperscript{\dag}}$ \\ 
\affiliations
$^1$National Key Laboratory for Novel Software Technology, Nanjing University, China\\
$^2$School of Artificial Intelligence, Nanjing University, China\\
$^3$School of Intelligence Science and Technology, Nanjing University, China\\
 \emails
 \{yangxw, shaojj, guolz, zhangbw, zhouz, jialh, daiwz, liyf\}@lamda.nju.edu.cn
}

\begin{document}

\maketitle

\begin{abstract}
Large Language Models (LLMs) have shown promising results across various tasks, yet their reasoning capabilities remain a fundamental challenge. Developing AI systems with strong reasoning capabilities is regarded as a crucial milestone in the pursuit of Artificial General Intelligence (AGI) and has garnered considerable attention from both academia and industry. Various techniques have been explored to enhance the reasoning capabilities of LLMs, with neuro-symbolic approaches being a particularly promising way. This paper comprehensively reviews recent developments in neuro-symbolic approaches for enhancing LLM reasoning. We first present a formalization of reasoning tasks and give a brief introduction to the neuro-symbolic learning paradigm. Then, we discuss neuro-symbolic methods for improving the reasoning capabilities of LLMs from three perspectives: \emph{Symbolic$\to$LLM}, \emph{LLM$\to$Symbolic}, and \emph{LLM$+$Symbolic}. Finally, we discuss several key challenges and promising future directions. We have also released a GitHub repository including papers and resources related to this survey: \href{https://github.com/LAMDASZ-ML/Awesome-LLM-Reasoning-with-NeSy}{{https://github.com/LAMDASZ-ML/Awesome-LLM-Reasoning-with-NeSy}}.
\end{abstract}

\section{Introduction}
The development of artificial intelligence (AI) has evolved through distinct phases, shaped by different paradigms. \emph{Symbolic AI}, focused on manipulating symbols, logic, rules, and knowledge to mimic human problem-solving abilities, laid the foundation for AI research before the 1990s. However, it encountered significant challenges, particularly regarding scalability and flexibility to real-world applications with noisy raw sensory inputs. On the other hand, \emph{Connectionist AI}, centered on neural networks and achieved remarkable success in data-driven machine learning. More recently, building on the success of transformer models, LLMs have demonstrated promising results in various tasks. However, many researchers have reported that LLMs struggle with complex reasoning problems; they only attempt to replicate reasoning steps in training data, and cannot really reason. More efforts must be devoted to overcoming these bottlenecks for developing strong reasoning models.

Building AI models with strong reasoning capabilities is a crucial milestone toward achieving AGI. To this end, numerous researchers have focused on enhancing the reasoning abilities of LLMs. Existing studies can be categorized into three categories based on the different stages of the reasoning model construction: \emph{Data Construction}, including how to automatically generate/augment/annotate/select data with reasoning paths; \emph{Fine-Tuning}, including supervised fine-tuning and reinforcement fine-tuning on reasoning specialized datasets, and \emph{Inference}, including inference techniques ranging from CoT to test-time scaling. Various large reasoning models have also been released, including OpenAI O1, Qwen-QwQ, DeepSeek-R1, etc. 

Among these explorations, Neuro-Symbolic (NeSy) methods demonstrate superior performance. NeSy aims to integrate the strengths of symbolic AI, which excels in complex reasoning, with neural networks, which are adept at learning from large datasets \cite{nesy3}. By integrating these approaches, we can build AI systems that not only learn from large datasets but also handle complex reasoning tasks in a human-like manner. NeSy AI aligns with the Dual Process Theory in cognitive science, which posits that human cognition consists of two systems: System 1, which is fast, intuitive, and unconscious (neural networks), and System 2, which is slower, more deliberate, and conscious, focusing on logical reasoning and problem-solving (symbolic reasoning). Therefore, NeSy is naturally a promising way to improve the reasoning abilities of LLMs.

In this paper, we aim to give a brief introduction on how to exploit NeSy methods to improve LLM reasoning. We first give a formulation of reasoning tasks in LLMs (\textsection \ref{reasoning}). Next, we give an introduction about the basic NeSy paradigm (\textsection \ref{nesy}). Then, we discuss how NeSy methods could improve the reasoning abilities of LLMs from three perspectives, \emph{Symbolic$\to$LLM}, \emph{LLM$\to$Symbolic}, \emph{LLM$+$Symbolic} (\textsection \ref{nesyreason}). Finally, by examining these advancements, we discuss open challenges in this field and outline potential future work (\textsection \ref{future}).

\section{What is Reasoning?}
\label{reasoning}
    
The general reasoning task can be characterized as a recursive process in which each step builds upon the preceding reasoning steps. Formally, the task is defined as follows: given an input problem $Q$ and background knowledge $K$, the goal is to get the answer $A$, satisfying $A = f(Q, K)$, where $f$ denotes the function mapping the problem and background knowledge to the answer. The $Q$ and $K$ can be represented in different forms, such as natural language or symbolic. For the reasoning tasks, we also have a reasoning path $Z$ consists of a sequence of intermediate reasoning steps, $Z = \{z_1, z_2, \ldots, z_n\}$, where $z_i$ represents the $i$-th step and satisfies $z_i = g_i(Q, K,z_{1},z_2,...,z_{i-1})$. Here, $g_i$ is a reasoning function of the $i$-th step that incorporates the input problem, the background knowledge, and the previous steps. The final answer $A$ corresponds to the result of the last step in the reasoning path, i.e., $A = z_n$.

The reasoning function $g(\cdot)$ plays a central role across various reasoning scenarios, determining how new reasoning steps or conclusions are generated based on background knowledge and prior reasoning steps. The specific interpretations of the reasoning function $g(\cdot)$ vary across three primary reasoning types: deductive reasoning, inductive reasoning, and abductive reasoning.

\begin{itemize}[left=0pt]
\item \textbf{Deductive Reasoning}: In deductive reasoning, the reasoning function $g(\cdot)$ applies rules in the knowledge base $K$ to the intermediate steps to generate new reasoning steps or conclusions. Formally, 
\begin{equation*}
    z_i = \texttt{ApplyLogicRules}(Q,K,z_{1},z_{2},...,z_{i-1})
\end{equation*}
where $\texttt{ApplyLogicRules}(\cdot)$ is an operation based on formal logic. 

\item \textbf{Inductive Reasoning}: Inductive reasoning involves generalizing patterns or rules from specific examples. The reasoning function $g(\cdot)$ extracts broader patterns from the intermediate steps. Formally,
\begin{equation*}
    z_i = \texttt{InducePattern}(Q,K,z_{1},z_{2},...,z_{i-1})
\end{equation*}
where $\texttt{InducePattern}(\cdot)$ is a pattern-discovery function, often relying on statistics, machine learning, or human expertise. 

\item \textbf{Abductive Reasoning}: Abductive reasoning generates hypotheses to explain observed phenomena. The reasoning function $g(\cdot)$ identifies the most possible hypothesis for the previous reasoning steps. Formally,
\begin{equation*}
    z_i = \texttt{GenerateHypothesis}(Q,K,z_{1},z_{2},...,z_{i-1})
\end{equation*}
where $\texttt{GenerateHypothesis}(\cdot)$ is a hypothesis generation function that seeks to identify the most plausible hypothesis to explain $z_{i-1}$. 
\end{itemize}

\section{What is Neuro-Symbolic AI?}
\label{nesy}

Neuro-symbolic AI seeks to combine the learning capabilities of neural networks with the reasoning power of symbolic AI. This integration enables the development of AI systems that can learn from extensive datasets while applying knowledge, rules, and logical reasoning, allowing them to tackle tasks that require both intuitive and deliberate thinking. Specifically, we categorize neuro-symbolic AI into three types: \emph{Neuro helps Symbolic}, \emph{Symbolic helps Neuro}, and \emph{Hybrid Neuro-Symbolic Architecture}. For a more fine-grained categorization, please refer to Henry A. Kautz's lecture in AAAI 2020~\cite{kautz2022third}.

\subsection{Neuro helps Symbolic}
\label{l4r}

This branch predominantly relies on symbolic processing, yet incorporates neural networks to address limitations of symbolic systems. Pure symbolic AI faces some key limitations, e.g., 1) large search spaces that hinder efficient problem-solving; 2) dependence on precisely defined symbols, limiting the representation of abstract real-world concepts; and 3) a rigid, deterministic reasoning process that struggles with ambiguity and uncertainty. These challenges can be addressed by integrating neural components, leading to the proposal of various NeSy approaches.

Exploiting neural networks to accelerate reasoning in symbolic systems represents a classic technique, with a notable example being AlphaGo \cite{alphago}. AlphaGo integrates neural networks into MCTS, employing reinforcement learning on large-scale datasets to train policy and value networks. This integration provides heuristic acceleration for symbolic search, significantly overcoming the limitation of the pure symbolic system. 

To overcome the limitations of symbolic systems in grounding abstract symbolic concepts into sub-symbolic real-world representations, neural networks can be exploited to directly extract symbolic concepts from raw data. This is a well-established and enduring research area~\cite{TaddeoF05}. One recent representative work is NS-CL \cite{mao2019neuro}, which uses a convolutional neural network to parse images and extract features, while symbolic reasoning modules interpret these features to answer complex questions about the images.

To address the limitations of symbolic systems in handling ambiguity and uncertain reasoning, various studies have focused on making the symbolic process differentiable. A prime example is $\partial$ILP \cite{phiilp}, which introduces the differentiable properties of neural networks into inductive logic programming (ILP). These approaches preserve symbolic logical reasoning while leveraging neural networks to process uncertain and probabilistic information, enabling the recognition of complex patterns.

\begin{figure*}[t]
    \centering
    \includegraphics[width=\linewidth]{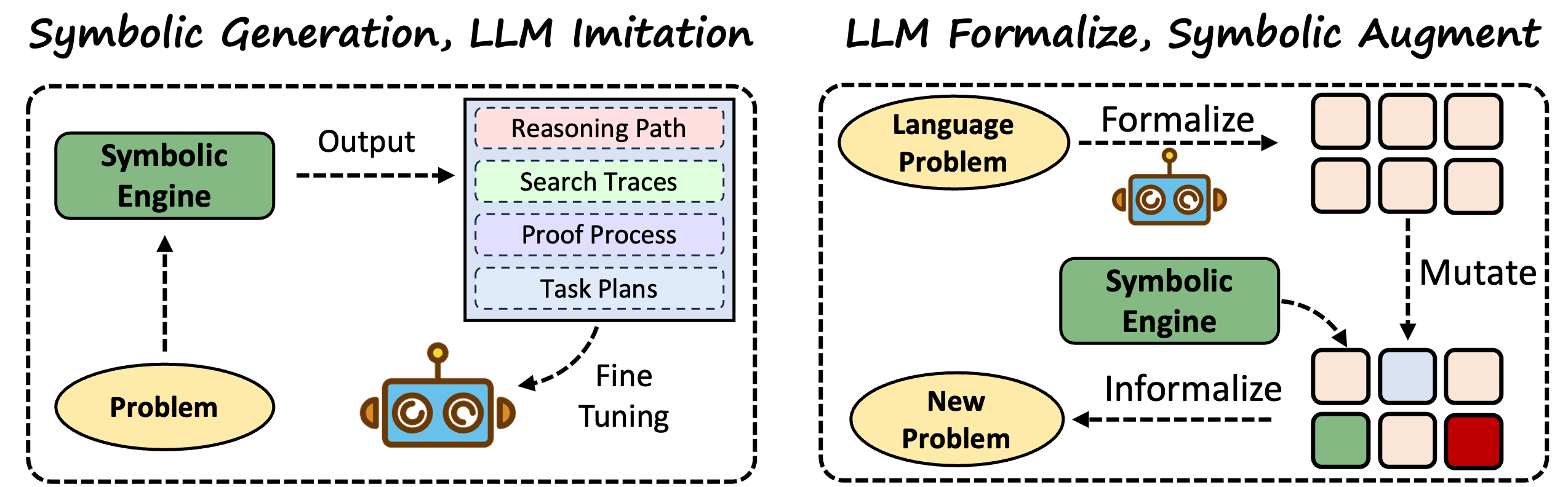}
    \caption{Illustration of how symbolic methods can be exploited to provide reasoning data for LLMs.}
    \label{syb4llm}
\end{figure*}

\subsection{Symbolic helps Neuro}
\label{r4l}
This branch of studies predominantly relies on neural networks, yet incorporates symbolic components to facilitate specialized features, such as logical reasoning, or improve interpretability and trustworthiness. Predominantly, there are two main methodologies to achieve this integration.

The first way is symbolic regularization approaches, which treat symbolic knowledge and rules as optimization constraints in the learning objective to guide the training process of neural networks, ensuring that the model’s predictions align more closely with domain-specific knowledge. Representative methods include Semantic Loss \cite{SL}, which directly incorporates logical constraints into the training process as a penalization term, and Logic Tensor Networks (LTNs) \cite{ltn}, which combine first-order logic with tensor-based computations, allowing neural networks to learn representations that satisfy logical rules.

The second way is model-based approaches, which try to modify the neural network by infusing symbolic knowledge into its structure directly, rather than externally guiding the training process of neural networks using constraints. This can be achieved by designing specific layers or modules that are informed by symbolic rules~\cite{RNM}. Compared with regularization techniques, model-based methods embed symbolic knowledge directly into the model’s structure and ensure the knowledge is always considered in both the training and inference stages.

\subsection{Hybrid Neuro-Symbolic Architecture}
\label{lri}
Different from previous categories that are ``symbolic-heavy" or ``neural-heavy", various studies seek to design new hybrid neuro-symbolic architecture to make symbolic and neural networks work simultaneously in a framework to better exploit their strengths.

DeepProbLog \cite{dpl} and Abductive Learning (ABL) \cite{abl} are two representative methods. DeepProbLog \cite{dpl} integrates deep learning and probabilistic logic, focusing on the interaction between learning and reasoning. It extends probabilistic programming by introducing neural predicates, which act as a bridge between neural networks and symbolic reasoning. This allows the system to harness the strengths of both paradigms: the pattern recognition capabilities of deep learning and the structured reasoning of logic programming. By enabling end-to-end training, DeepProbLog demonstrates powerful abilities for tasks requiring both perception and logical inference. Abductive learning \cite{abl} offers a framework that bridges machine learning and logical reasoning through inconsistency minimization, enabling the generation of pseudo-labels for intermediate symbolic concepts. Unlike DeepProbLog, ABL does not aim to make the symbolic system differentiable. Instead, it fully leverages the reasoning capabilities inherent in symbolic knowledge by employing abductive reasoning to sample pseudo-labels for intermediate symbols. Both methods are capable of concurrently updating neural networks and symbolic systems.

In this paper, we follow the above categorization and discuss neuro-symbolic AI for improving the reasoning abilities of LLMs in a similar categorization, i.e., Symbolic$\to$LLM, LLM$\to$Symbolic, and LLM$+$Symbolic. Given the breadth of research in this area, we respectfully acknowledge that our discussion is limited to a subset of representative works that most effectively convey the key concepts.

\section{Symbolic $\to$ LLM: Addressing the Reasoning Data Scarcity}
\label{nesyreason}
Creating large-scale reasoning datasets with high-quality reasoning paths is essential for enhancing the reasoning capabilities of LLMs. However, constructing such datasets poses significant challenges, as ensuring logical rigor and coherence in step-by-step reasoning processes is inherently difficult. Moreover, data annotation, particularly step-wise annotation, is highly resource-intensive, further complicating the dataset development process. In contrast, symbolic methods are characterized by their rigorous reasoning abilities, offering a promising way to investigate how they might compensate for the lack of sufficient reasoning data.

In this section, we illustrate how to explore symbolic methods to help address the reasoning data scarcity problem. Specifically, we explore this direction from two perspectives, \emph{Symbolic Generation, LLM Imitation} and \emph{LLM Formalize, Symbolic Augment}. The main ideas of these paradigms are illustrated in Figure~\ref{syb4llm}.

\begin{figure*}[t]
    \centering
    \includegraphics[width=\linewidth]{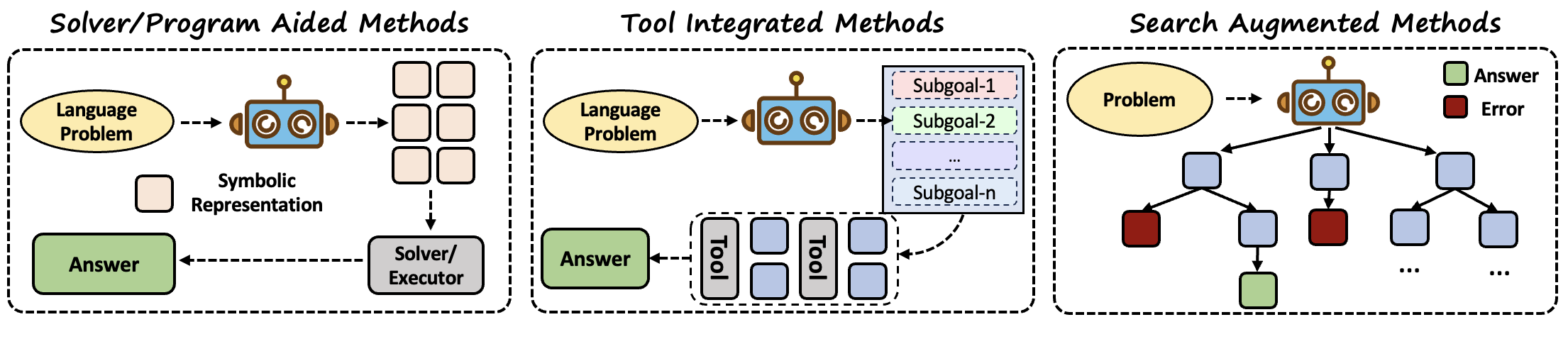}
    \caption{Illustration of how LLMs integrate symbolic solvers, programs, tools, or search algorithms to facilitate the reasoning process.}
    \label{llm4sym}
\end{figure*}

\subsection{Symbolic Generation, LLM Imitation} 

For problems that can be solved using symbolic methods, we can leverage these methods to generate logically rigorous reasoning paths. Fine-tuning LLMs on such datasets enables them to learn and potentially mimic the reasoning capabilities of symbolic methods. This idea is quite similar to knowledge distillation, but instead of distilling data from a stronger model, it distills data from symbolic methods. Representative symbolic methods that can be applied to generate reasoning data, including but not limited to logical reasoning algorithms, logic solvers, constraint optimization, search algorithms, etc.

Numerous efforts have been devoted to leveraging various symbolic methods for generating reasoning data and fine-tuning LLMs to be applied in different tasks. For example, AlphaGeometry~\cite{alphageometry} develops a symbolic deduction engine to obtain the reasoning solutions to geometry problems, and surpasses the performance of the average IMO contestant in proving Euclidean plane geometry theorems. LOGIPT~\cite{LOGIPT} proposes to generate logical reasoning processes via logical solvers and fine-tuning LLMs to mimic the reasoning process of logical solvers, enabling LLMs to acquire similar abilities for tackling deductive reasoning tasks. Procedure Cloning~\cite{yang2022chain}, DualFormer~\cite{dualformer}, and SOS~\cite{SOS} generate search traces with search algorithms, such as DFS, BFS, A$^*$, MCTS, etc, and fine-tune LLMs to enable the model to learn to search and backtrack in the reasoning procedure. Planformer~\cite{plansformer} generates plans for classical planning tasks with the FastDownward planner and constructs a PDDL-based dataset to fine-tune LLMs for planning tasks. The goal of these methods is to internalize the abilities of the symbolic solvers into the LLMs by constructing datasets for fine-tuning the model, thereby enhancing the LLM’s reasoning abilities.

\subsection{LLM Formalize, Symbolic Augment}
Unlike previous studies that focus on generating reasoning data with symbolic methods and fine-tuning LLMs to mimic their capabilities, this line of research takes a different direction that aims to augment data with symbolic methods. The pipeline can typically be summarized as: firstly, we can transform informal natural language data into a formalized representation space; secondly, we employ symbolic rules or solvers to help augment the data in the formalized space; thirdly, we transform the symbolically represented data into natural language. The help of symbolic rules or solvers enables the automatic augmentation of reasoning data that is both linguistically diverse and logically rigorous, compared to data augmentation with LLMs solely.

There are also various studies in this direction. For example, NSDG~\cite{nsdg} formalizes the language represented math problems into its symbolic version, represented by the SMT-LIB language, and mutates the symbolic problem to create new variants for the data augmentation. Then, the symbolic form is converted to the natural language version. LLMs act as a bridge between symbolic and natural language spaces in this process. AMR-DA~\cite{AMR-DA} converts the original language data into an abstract meaning representation graph (AMR), a structured semantic representation that encapsulates the logical structure of the sentence, upon which operations are performed to generate logically modified AMR graphs. The modified AMR graphs are subsequently converted back into text to create augmented data. Similar ideas have also been adopted in the legal reasoning~\cite{zhou2025lawgpt}, logical reasoning~\cite{qi2025large}, theorem proving~\cite{alphaintegrator}, etc. Such methods excel in ensuring the logical correctness of generated data through symbolic reasoning and unlocking the potential for unlimited data generation, as the complexity of symbolic spaces provides nearly infinite possibilities, greatly expanding the diversity and scale of available reasoning datasets.

\section{LLM $\to$ Symbolic: Addressing the Reasoning Function Error}

From the formulation of reasoning tasks, it can be observed that the reasoning function $g(\cdot)$ serves as the core component in the reasoning process. LLMs typically employ auto-regressive techniques to approximate this reasoning function. However, this approach inherently introduces errors. Even minor inaccuracies at each step can propagate and amplify over successive steps, eventually causing the reasoning outcomes to deviate substantially from the correct answers.

To address errors in reasoning functions, it is a promising way to replace the auto-regressive style reasoning function with external symbolic methods to enhance the rigor of intermediate reasoning steps. The underlying intuition is that LLMs are inherently less adept at precise, long-chain reasoning. Therefore, enabling them to learn how to invoke external modules provides a more effective approach to solving complex reasoning problems. Formally, for given indices $i$ and $j$ ($j>i$), this type of method can be expressed as: $z_j = \texttt{ExternalModules}(z_i)$. Here, the reasoning steps are represented symbolically, serving as a prerequisite for applying such methods. The \texttt{ExternalModules} represents an external module that processes the input $z_i$ and produces the output $z_j$ directly. The commonly external modules include symbolic solvers, program interpreters, off-the-shelf models, APIs, tools, search algorithms, etc. The main ideas of these methods are illustrated in Figure~\ref{llm4sym}.

\subsection{Symbolic Solver Aided Methods}
The core idea of symbolic solver aided methods is to utilize external symbolic solvers, such as logic programming, SMT solvers, theorem provers, PDDL planners, constraint optimization tools, etc, to replace multiple reasoning functions performed internally. The pipeline of these methods is straightforward: first, translate natural language problems into a formalized representation that symbolic solvers can process, then call the solver to derive the solution.

LogicLM~\cite{logiclm} and LINC~\cite{LINC} are some of the earlier attempts at this approach. These methods transform natural language problems into executable logical expressions and leverage logic solvers to obtain results. There is also a branch of studies that focuses on exploiting classical planning methods to deal with planning tasks such as classical planning or robotics. For example, LLM+P~\cite{LLM+P} propose to incorporate a PPDL-based symbolic planner. It leverages the semantic understanding abilities of LLM to translate informal language into a formal PDDL language and then employs the Fast-Downward solver for the planning process.

An important research direction related to these methods is how to automatically translate natural language represented data into a formalized symbolic representation. This is also called \emph{AutoFormalization}. Representative studies include autoformalization of first-order logic, mathematical statements, mathematical proofs, PDDL, symbolic world models, etc. Key challenges lie in how to improve the consistency and the efficiency of the autoformalization process.

\subsection{Program-Aided Methods}
Instead of exploiting symbolic solvers, there are also various studies that aim to exploit the program interpreter to help improve the accuracy of reasoning functions. The main pipeline is similar to symbolic solver aided methods, with the key difference being that these methods convert natural language into a programming language and leverage a program executor to derive the solution.

PAL (Program-Aided Language Model)~\cite{pal} and PoT (Program of Thought)~\cite{pot} are two representative program-aided methods. Specifically, they adopt LLMs to express the reasoning process as Python programming languages, and the computation is relegated to an external program executor. Except for the Python programs, Binder~\cite{binding} converts the input problem to SQL and executes the program interpreter to obtain the answer. Instead of relying on the code interpreter solely, CoC (Chain of Code)~\cite{coc} generates code or pseudo-code and then executes the code with a code interpreter if possible, and with an LMulator (language model emulating code) otherwise. Related methods have been successfully applied to mathematical reasoning, code generation, robotics, etc. These methods can be seen as augmenting the chain of thought with external program interpreters to enable more accurate and robust reasoning. 

\subsection{Tool-Aided Methods}
Beyond symbolic solvers and program interpreters, numerous other tools, APIs, and off-the-shelf models that have been developed for various tasks can also be exploited to enhance the accuracy of reasoning functions. Examples include calculators for numerical computations, web search engines for common-sense reasoning, and pre-trained vision models for visual reasoning.

Different from symbolic solver or program aided methods, which translate the natural language problems into a formalized language, and call the symbolic solver or program executor to obtain the solution directly, tool-aided methods are more complex, since they typically require calling different tools in different reasoning steps. Therefore, the pipeline can be divided into four stages: \emph{task planning}, \emph{tool selection}, \emph{tool calling}, \emph{response generation}. 

For example, in the visual reasoning tasks, VisProg~\cite{visprog} utilizes LLMs to generate a Python-like API call program, integrating various tools, such as image processing subroutines in OpenCV, and off-the-shelf vision models, to perform complex visual reasoning tasks based on natural language instructions. These ideas inspired various subsequent studies such as ViperGPT~\cite{vipergpt}, Chameleon~\cite{chameleon}, VisualSkechpad~\cite{huvisual}, etc. For mathematical reasoning, Tora~\cite{tora} explores the integration of LLMs with the utilization of external tools such as computation libraries and symbolic solvers. The key differences among these methods lie in the choice of tool libraries tailored to specific tasks and the strategies used to equip LLMs with accurate tool-utilization capabilities. These strategies include prompt techniques and fine-tuning the model on a diverse set of collected tool-use trajectories, such as SFT or RLFT.

\begin{figure*}[t]
    \centering
    \includegraphics[width=\linewidth]{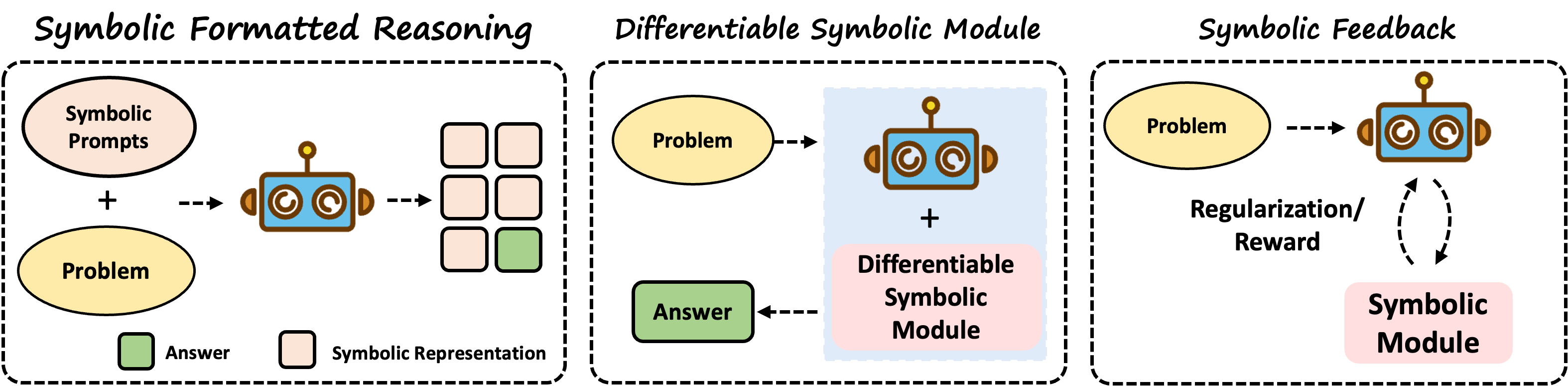}
    \caption{Illustration of the main ideas of end-to-end LLM+symbolic reasoning, including symbolic formatted reasoning, differential symbolic modules, and symbolic feedback.}
    \label{end2end}
\end{figure*}

\subsection{Search Augmented Methods}
The core process of traditional symbolic solvers, such as SMT solvers or PDDL planners, can be abstractly modeled as a \textit{search} problem: finding solutions that satisfy given constraints within a potentially vast or even infinite solution space, and adopting various pruning strategies and optimization techniques to accelerate the search process. To achieve similar abilities to these symbolic methods, various methods aim to directly combine the search algorithms with the decoding procedure of LLMs to enhance the reasoning capabilities of LLMs. Formally, at the $i$-th step of reasoning, $T$ reasoning functions are expanded, producing $T$ candidate reasoning results: $z_i^{(t)} = g_i^{(t)}(Q, K, z_1, z_2, \ldots, z_{i-1}),t=1,2,...,T$, where $g_i^{(t)}$ represents the $t$-th reasoning function, and this can be realized by using the sampling mechanism of a language model. Consequently, the reasoning process can be transformed into a search problem, allowing the incorporation of various search algorithms.

For example, DBS~\cite{zhudeductive} proposes a decoding algorithm that integrates self-evaluation guidance via the beam search. SPaR~\cite{spar} proposes a self-play framework integrating BFS and DFS to refine the response and improve the instruction-following abilities of LLMs. ChinaTravel~\cite{shao2024chinatravel} exploits BFS and DFS to improve the travel planning abilities of LLMs. NeurologicA*~\cite{neurologicA*} incorporates heuristic estimates of future cost in the decoding process, like the A* search algorithms. Moreover, building on the advantages of MCTS in AlphaGo and AlphaZero, numerous studies have been devoted to exploring the integration of MCTS in the LLMs' inference stage to extend reasoning chains and improve reasoning accuracy. The difference in this line of research mainly lies in the choice of search algorithms and their intended role, whether to guide more effective exploration in the reasoning process or to refine reward estimation through simulation. However, these methods have clear limitations, since more exploration leads to a huge computation cost in the inference time.

\section{Symbolic $+$ LLMs: End-to-End Reasoning}
In the previous two branches of research, LLMs and symbolic methods operate separately. For example, for the symbolic$\to$LLMs paradigm, symbolic AI methods are responsible for generating or augmenting data, after which the LLMs are fine-tuned on these datasets. For LLMs$\to$symbolic paradigm, the LLMs call symbolic solvers, which then derive the answer. A holy-grail problem for neuro-symbolic AI  is designing mechanisms that enable symbolic methods and neural networks to work in a more hybrid or end-to-end fashion. Specifically, we review efforts about Symbolic$+$LLMs in this direction from three perspectives, i.e., symbolic formatted reasoning, differential symbolic module, and symbolic feedback. The main ideas are illustrated in Figure~\ref{end2end}.

\subsection{Symbolic Formatted Reasoning}
For the reasoning process in LLMs, the reasoning path $Z$ can be represented in various forms, such as natural language or latent embedding. In certain reasoning tasks, these representations may be prone to inaccuracies. Such representational errors can accumulate progressively as the reasoning chain lengthens, ultimately leading to significant deviations from the correct reasoning solutions. 

To mitigate potential representation errors in reasoning chains, we may adopt formally symbolic representations to describe intermediate reasoning states, instead of relying solely on natural language descriptions or latent embedding. This symbolic representation offers significant advantages by providing more explicit and precise semantic expressions, thereby avoiding errors caused by the ambiguity of natural language or the vagueness of latent embedding. Formally, $\forall i \neq n, z_i \in \mathcal{L}$, where $\mathcal{L}$ denotes a symbolic language defined over a predetermined alphabet. The final answer, $A$, is derived from this sequence of intermediate symbolic representations. 

Symbolic formatted reasoning is particularly suitable for tasks that can be precisely symbolized, such as mathematical reasoning. For example, NaturalPrompt~\cite{ling2023deductive} enables LLMs to generate deductive reasoning chains using a proposed natural program format, Chain-of-Symbol prompting~\cite{cos} prompts LLMs to convert natural language into concise symbolic representations, enhancing their performance on spatial tasks while significantly reducing token consumption. LogicGuide~\cite{PoesiaGZG24} introduces a general logical reasoning system to assist the LLM. It formalizes the reasoning process within LogicGuide, ensuring that step-by-step reasoning remains sound and reliable. There are also some methods that use programming languages like Python as the symbolic language for intermediate representations \cite{weir2024learning}. Unlike program-aided methods, the Python code is not executed but rather serves solely as a structured prompt to guide the model toward the final answer. These methods leverage symbolic representations to mitigate potential representation error accumulation inherent in natural language representations during the reasoning process. Alongside natural language and latent embedding representation reasoning, they form the three primary reasoning formations.

\subsection{Differential Symbolic Module}
Unlike machine learning, which relies on continuous numerical optimization, symbolic AI methods are typically grounded in discrete symbolic reduction. Consequently, designing a unified optimization framework that allows machine learning and symbolic reasoning to be jointly optimized remains a fundamental challenge. A promising direction is to develop differential symbolic modules to enable the symbolic reasoning systems to work with the machine learning models. 

In classical NeSy studies, many efforts have been devoted to this direction. More recently, studies have emerged to develop differentiable symbolic modules that can seamlessly integrate with LLMs. For example, DiLA~\cite{zhang2024dila} leverages LLMs to parse natural language represented problem into a SAT problem which consists of a set of logical formulas, then LLM generates an initial solution based on its natural language understanding, next the relaxed variable and extracted constraints are offloaded to a differential logic layer, checking the constraint satisfiability, and updating the solution until all constraints are met. Oreoml~\cite{oreoml} propose to integrate knowledge graph reasoning of symbolic logic and neural networks, enabling the LLMs to work together with a differentiable knowledge graph reasoning module through the knowledge interaction layers embedded in the LLMs. NSVQA~\cite{NSVQA} leverages predicates as symbolic representations linked to neural modules that process visual elements, disentangle visual representation learning from the inference mechanism, and propose a differentiable first-order logic formalism based on fuzzy logic for compositional visual reasoning. AutoCoNN~\cite{AutoCoNN} proposes a framework that synergistically integrates compiled neural networks (CoNNs) into the standard transformer architecture. CoNNs are neural modules designed to explicitly encode rules through artificially generated attention weights. By incorporating CoNN modules, the neural comprehension framework enables LLMs to execute rule-intensive symbolic tasks. 

These methods represent initial attempts to enable joint optimization of logical reasoning and LLM fine-tuning. However, they still face significant challenges, such as open concepts, optimization efficiency, generality, etc. More efforts are needed to improve the efficiency, scalability, and robustness of these methods.

\subsection{Symbolic Feedback}
Supervised fine-tuning and reinforcement fine-tuning are two main learning paradigms for enhancing the reasoning capabilities of pre-trained LLMs. Both rely on supervisory signals, such as the supervised loss and the reward function, to guide the learning process. Symbolic methods offer more precise and interpretable verification capabilities. Therefore, exploring how to leverage them as regularization terms in the loss function for supervised fine-tuning or reward signals in reinforcement fine-tuning represents a promising and important research direction.

For supervised fine-tuning, \cite{premsri2024neuro} proposes to fine-tune LLMs by exploiting the spatial logical rules as constraints, providing additional supervision to improve spatial reasoning. For reinforcement fine-tuning, SyreLM~\cite{syrelm} adopts a small frozen LM, equipped with an adapter,  to translate natural language problems into formal language expressions, and adopts reinforcement learning to train the adapted LM, informed by the non-differentiable symbolic solver. RBR~\cite{murule} proposes rule-based rewards to improve the safety of LLMs. RLSF~\cite{jha2024rlsf} exploits various reasoning or domain knowledge tools (e.g., symbolic solvers, theorem provers, or knowledge bases) to provide feedback to the LLMs. Similar ideas have also been adopted by LLM-Modulo~\cite{kambhampati2024position} for planning tasks, Cotran~\cite{cotran} for code generation tasks, etc.  This line of research is related to neuro-symbolic reinforcement learning. The advantages of these methods lie in that it does not require the symbolic methods to be differentiable.

\section{Challenges \& Open Research Directions}
\label{future}
Despite the promising advances in NeSy approaches for enhancing reasoning capabilities in LLMs, several challenges persist. This section discusses key challenges and proposes open research directions to address them.

\paragraph{Multi-Modal Reasoning} Previous studies mainly focus on reasoning in language modality. Various real-world applications require multi-modal reasoning, such as VQA, embodied AI, spatial intelligence, etc. How to effectively exploit symbolic systems that are integrated with multi-modal language models remains an open problem. Moreover, existing multi-modal reasoning is mainly conducted on the language modalities. In contrast, human reasoning often involves the simultaneous processing of multiple modalities. For example, when solving geometry problems, humans may draw auxiliary lines on diagrams to support their reasoning. This highlights a significant gap between current approaches and human manner multi-modal reasoning.

\paragraph{Advanced Hybrid Architectures} Though LLMs achieve promising results across various tasks, they remain data-driven machine learning models that rely on statistical pattern recognition rather than formal logical reasoning. To construct AI systems with both abilities to exploit big data and perform rigorous reasoning, it is important to develop more advanced hybrid neuro-symbolic architectures that seamlessly integrate LLMs and symbolic reasoning components. More advanced optimization techniques are also required to improve the scalability and efficiency of these architectures.

\paragraph{Theoretical Understanding} The theoretical understanding of how symbolic methods enhance the reasoning abilities of LLMs is crucial for guiding the design of more effective algorithms. However, efforts towards this direction remain limited. For example, the generalization performance with symbolic methods, optimization theory for the LLMs fine-tuning with symbolic feedback, analysis on the reasoning shortcuts, the relationship between scaling laws and the integration of symbolic methods, etc. Further efforts are needed to establish solid theoretical foundations for LLMs' reasoning.

\section{Conclusion}
\label{conclusion}

Improving the reasoning capabilities of AI models is a key milestone towards AGI. This paper explores the role of neuro-symbolic methods in enhancing reasoning in LLMs. We give a formulation of reasoning tasks and introduce the main ideas from three perspectives, Symbolic $\to$ LLM, LLM $\to$ Symbolic, and LLM $+$ Symbolic, to explore how symbolic methods can be exploited to address critical challenges in LLMs’ reasoning, including reasoning data scarcity, reasoning function errors, representation errors, etc. Open challenges and opportunities are also discussed. 

Given the rapid development of this field and the vast number of publications, it is not feasible to cover all papers comprehensively. Instead, we focus on presenting the key ideas and technical approaches. We aim for this paper to provide an up-to-date summary of recent advancements and inspire new insights into the integration of symbolic AI methods with LLMs for the development of strong reasoning models.

% \appendix

% \section*{Ethical Statement}

% \section*{Acknowledgments}

% \newpage

\section*{Acknowledgments}
This research was supported by National Science Foundation of China (62306133, 62206124), Key Program of Jiangsu Science Foundation (BK20243012), Jiangsu Science Foundation Leading-edge Technology Program (BK20232003) and Major Program (JD) of Hubei Province (2023BAA024).

\section*{Contribution Statement}
Equal contribution ($*$): Xiao-Wen Yang, Jie-Jing Shao and Lan-Zhe Guo. Corresponding author($\dag$): Yu-Feng Li.

\bibliographystyle{named}
\bibliography{ijcai25}

\begin{thebibliography}{}

\bibitem[\protect\citeauthoryear{Amizadeh \bgroup \em et al.\egroup }{2020}]{NSVQA}
Saeed Amizadeh, Hamid Palangi, Alex Polozov, Yichen Huang, and Kazuhito Koishida.
\newblock Neuro-symbolic visual reasoning: Disentangling visual from reasoning.
\newblock In {\em ICML}, pages 279--290, 2020.

\bibitem[\protect\citeauthoryear{Badreddine \bgroup \em et al.\egroup }{2022}]{ltn}
Samy Badreddine, Artur~d'Avila Garcez, Luciano Serafini, and Michael Spranger.
\newblock Logic tensor networks.
\newblock {\em Artificial Intelligence}, 2022.

\bibitem[\protect\citeauthoryear{Chen \bgroup \em et al.\egroup }{2023}]{pot}
Wenhu Chen, Xueguang Ma, Xinyi Wang, and William~W Cohen.
\newblock Program of thoughts prompting: Disentangling computation from reasoning for numerical reasoning tasks.
\newblock {\em TMLR}, 2023.

\bibitem[\protect\citeauthoryear{Cheng \bgroup \em et al.\egroup }{2023}]{binding}
Zhoujun Cheng, Tianbao Xie, Peng Shi, et~al.
\newblock Binding language models in symbolic languages.
\newblock In {\em ICLR}, 2023.

\bibitem[\protect\citeauthoryear{Cheng \bgroup \em et al.\egroup }{2025}]{spar}
Jiale Cheng, Xiao Liu, Cunxiang Wang, Xiaotao Gu, Yida Lu, Dan Zhang, Yuxiao Dong, Jie Tang, Hongning Wang, and Minlie Huang.
\newblock Spar: Self-play with tree-search refinement to improve instruction-following in large language models.
\newblock In {\em ICLR}, 2025.

\bibitem[\protect\citeauthoryear{De~Raedt \bgroup \em et al.\egroup }{2020}]{nesy3}
Luc De~Raedt, Sebastijan Duman{{c}}i{'c}, Robin Manhaeve, and Giuseppe Marra.
\newblock From statistical relational to neuro-symbolic artificial intelligence.
\newblock {\em preprint arXiv:2003.08316}, 2020.

\bibitem[\protect\citeauthoryear{Dutta \bgroup \em et al.\egroup }{2024}]{syrelm}
Subhabrata Dutta, Ishan Pandey, Joykirat Singh, Sunny Manchanda, Soumen Chakrabarti, and Tanmoy Chakraborty.
\newblock Frugal {LM}s trained to invoke symbolic solvers achieve parameter-efficient arithmetic reasoning.
\newblock In {\em AAAI}, pages 17951--17959, 2024.

\bibitem[\protect\citeauthoryear{Evans and Grefenstette}{2018}]{phiilp}
Richard Evans and Edward Grefenstette.
\newblock Learning explanatory rules from noisy data.
\newblock {\em JAIR}, 2018.

\bibitem[\protect\citeauthoryear{Feng \bgroup \em et al.\egroup }{2024}]{LOGIPT}
Jiazhan Feng, Ruochen Xu, Junheng Hao, Hiteshi Sharma, Yelong Shen, Dongyan Zhao, and Weizhu Chen.
\newblock Language models can be deductive solvers.
\newblock In {\em NAACL Findings}, pages 4026--4042, 2024.

\bibitem[\protect\citeauthoryear{Gandhi \bgroup \em et al.\egroup }{2024}]{SOS}
Kanishk Gandhi, Denise~HJ Lee, Gabriel Grand, Muxin Liu, Winson Cheng, Archit Sharma, and Noah Goodman.
\newblock Stream of search ({SOS}): Learning to search in language.
\newblock In {\em COLM}, 2024.

\bibitem[\protect\citeauthoryear{Gao \bgroup \em et al.\egroup }{2023}]{pal}
Luyu Gao, Aman Madaan, Shuyan Zhou, Uri Alon, Pengfei Liu, Yiming Yang, Jamie Callan, and Graham Neubig.
\newblock {PAL}: Program-aided language models.
\newblock In {\em ICML}, pages 10764--10799, 2023.

\bibitem[\protect\citeauthoryear{Gou \bgroup \em et al.\egroup }{2024}]{tora}
Zhibin Gou, Zhihong Shao, Yeyun Gong, Yelong Shen, Yujiu Yang, Minlie Huang, Nan Duan, and Weizhu Chen.
\newblock Tora: A tool-integrated reasoning agent for mathematical problem solving.
\newblock In {\em ICLR}, 2024.

\bibitem[\protect\citeauthoryear{Gupta and Kembhavi}{2023}]{visprog}
Tanmay Gupta and Aniruddha Kembhavi.
\newblock Visual programming: Compositional visual reasoning without training.
\newblock In {\em CVPR}, 2023.

\bibitem[\protect\citeauthoryear{Hu \bgroup \em et al.\egroup }{2022}]{oreoml}
Ziniu Hu, Yichong Xu, Wenhao Yu, Shuohang Wang, Ziyi Yang, Chenguang Zhu, Kai-Wei Chang, and Yizhou Sun.
\newblock Empowering language models with knowledge graph reasoning for open-domain question answering.
\newblock In {\em EMNLP}, pages 9562--9581, 2022.

\bibitem[\protect\citeauthoryear{Hu \bgroup \em et al.\egroup }{2024a}]{cos}
Hanxu Hu, Hongyuan Lu, Huajian Zhang, Yun-Ze Song, Wai Lam, and Yue Zhang.
\newblock Chain-of-symbol prompting for spatial reasoning in large language models.
\newblock In {\em COLM}, 2024.

\bibitem[\protect\citeauthoryear{Hu \bgroup \em et al.\egroup }{2024b}]{huvisual}
Yushi Hu, Weijia Shi, Xingyu Fu, Dan Roth, Mari Ostendorf, Luke Zettlemoyer, Noah~A Smith, and Ranjay Krishna.
\newblock Visual sketchpad: Sketching as a visual chain of thought for multimodal language models.
\newblock In {\em NeurIPS}, 2024.

\bibitem[\protect\citeauthoryear{Jana \bgroup \em et al.\egroup }{2024}]{cotran}
Prithwish Jana, Piyush Jha, Haoyang Ju, Gautham Kishore, Aryan Mahajan, and Vijay Ganesh.
\newblock Cotran: An {LLM}-based code translator using reinforcement learning with feedback from compiler and symbolic execution.
\newblock In {\em ECAI}, pages 4011--4018, 2024.

\bibitem[\protect\citeauthoryear{Jha \bgroup \em et al.\egroup }{2024}]{jha2024rlsf}
Piyush Jha, Prithwish Jana, Pranavkrishna Suresh, Arnav Arora, and Vijay Ganesh.
\newblock {RLSF}: Reinforcement learning via symbolic feedback.
\newblock {\em preprint arXiv:2405.16661}, 2024.

\bibitem[\protect\citeauthoryear{Kambhampati \bgroup \em et al.\egroup }{2024}]{kambhampati2024position}
Subbarao Kambhampati, Karthik Valmeekam, Lin Guan, et~al.
\newblock Position: {LLMs} can't plan, but can help planning in {LLM-Modulo} frameworks.
\newblock In {\em ICML}, pages 22895--22907, 2024.

\bibitem[\protect\citeauthoryear{Kautz}{2022}]{kautz2022third}
Henry Kautz.
\newblock The third {AI} summer: {AAAI} robert s. engelmore memorial lecture.
\newblock {\em AI Magazine}, 43(1):105--125, 2022.

\bibitem[\protect\citeauthoryear{Li \bgroup \em et al.\egroup }{2023}]{coc}
Chengshu Li, Jacky Liang, Andy Zeng, Xinyun Chen, Karol Hausman, Dorsa Sadigh, Sergey Levine, Li~Fei-Fei, Fei Xia, and Brian Ichter.
\newblock Chain of code: Reasoning with a language model-augmented code emulator.
\newblock In {\em ICML}, pages 28259--28277, 2023.

\bibitem[\protect\citeauthoryear{Li \bgroup \em et al.\egroup }{2024}]{nsdg}
Zenan Li, Zhi Zhou, Yuan Yao, Yu-Feng Li, Chun Cao, Fan Yang, Xian Zhang, and Xiaoxing Ma.
\newblock Neuro-symbolic data generation for math reasoning.
\newblock In {\em NeurIPS}, pages 23488--23515, 2024.

\bibitem[\protect\citeauthoryear{Ling \bgroup \em et al.\egroup }{2023}]{ling2023deductive}
Zhan Ling, Yunhao Fang, Xuanlin Li, Zhiao Huang, Mingu Lee, Roland Memisevic, and Hao Su.
\newblock Deductive verification of chain-of-thought reasoning.
\newblock In {\em NeurIPS}, pages 36407--36433, 2023.

\bibitem[\protect\citeauthoryear{Liu \bgroup \em et al.\egroup }{2023}]{LLM+P}
Bo~Liu, Yuqian Jiang, Xiaohan Zhang, Qiang Liu, Shiqi Zhang, Joydeep Biswas, and Peter Stone.
\newblock {LLM+P}: Empowering large language models with optimal planning proficiency.
\newblock {\em preprint arXiv:2304.11477}, 2023.

\bibitem[\protect\citeauthoryear{Lu \bgroup \em et al.\egroup }{2022}]{neurologicA*}
Ximing Lu, Sean Welleck, Peter West, Liwei Jiang, Jungo Kasai, Daniel Khashabi, Ronan Le~Bras, Lianhui Qin, Youngjae Yu, Rowan Zellers, et~al.
\newblock Neurologic {A*} esque decoding: Constrained text generation with lookahead heuristics.
\newblock In {\em NAACL}, pages 780--799, 2022.

\bibitem[\protect\citeauthoryear{Lu \bgroup \em et al.\egroup }{2024}]{chameleon}
Pan Lu, Baolin Peng, Hao Cheng, Michel Galley, Kai-Wei Chang, Ying~Nian Wu, Song-Chun Zhu, and Jianfeng Gao.
\newblock Chameleon: Plug-and-play compositional reasoning with large language models.
\newblock In {\em NeurIPS}, pages 43447--43478, 2024.

\bibitem[\protect\citeauthoryear{Manhaeve \bgroup \em et al.\egroup }{2019}]{dpl}
Robin Manhaeve, Sebastijan Dumancic, Angelika Kimmig, Thomas Demeester, and Luc~De Raedt.
\newblock Deep{P}rob{L}og: Neural probabilistic logic programming.
\newblock {\em BNAIC}, 2019.

\bibitem[\protect\citeauthoryear{Mao \bgroup \em et al.\egroup }{2019}]{mao2019neuro}
Jiayuan Mao, Chuang Gan, Pushmeet Kohli, Joshua~B Tenenbaum, and Jiajun Wu.
\newblock The neuro-symbolic concept learner: Interpreting scenes, words, and sentences from natural supervision.
\newblock In {\em ICLR}, 2019.

\bibitem[\protect\citeauthoryear{Marra \bgroup \em et al.\egroup }{2020}]{RNM}
Giuseppe Marra, Michelangelo Diligenti, Francesco Giannini, Marco Gori, and Marco Maggini.
\newblock Relational neural machines.
\newblock In {\em ECAL}, 2020.

\bibitem[\protect\citeauthoryear{Mu \bgroup \em et al.\egroup }{2024}]{murule}
Tong Mu, Alec Helyar, Johannes Heidecke, Joshua Achiam, Andrea Vallone, Ian~D Kivlichan, Molly Lin, Alex Beutel, John Schulman, and Lilian Weng.
\newblock Rule based rewards for language model safety.
\newblock In {\em NeurIPS}, pages 108877--108901, 2024.

\bibitem[\protect\citeauthoryear{Olausson \bgroup \em et al.\egroup }{2023}]{LINC}
Theo Olausson, Alex Gu, Benjamin Lipkin, Cedegao~E. Zhang, Armando Solar{-}Lezama, Joshua~B. Tenenbaum, and Roger Levy.
\newblock {LINC:} {A} neurosymbolic approach for logical reasoning by combining language models with first-order logic provers.
\newblock In {\em EMNLP}, pages 5153--5176, 2023.

\bibitem[\protect\citeauthoryear{Pallagani \bgroup \em et al.\egroup }{2022}]{plansformer}
Vishal Pallagani, Bharath Muppasani, Keerthiram Murugesan, et~al.
\newblock Plansformer: Generating symbolic plans using transformers.
\newblock {\em preprint arXiv:2212.08681}, 2022.

\bibitem[\protect\citeauthoryear{Pan \bgroup \em et al.\egroup }{2023}]{logiclm}
Liangming Pan, Alon Albalak, Xinyi Wang, and William~Yang Wang.
\newblock Logic-{LM}: Empowering large language models with symbolic solvers for faithful logical reasoning.
\newblock In {\em EMNLP}, pages 3806--3824, 2023.

\bibitem[\protect\citeauthoryear{Poesia \bgroup \em et al.\egroup }{2024}]{PoesiaGZG24}
Gabriel Poesia, Kanishk Gandhi, Eric Zelikman, and Noah~D. Goodman.
\newblock Certified deductive reasoning with language models.
\newblock {\em TMLR}, 2024.

\bibitem[\protect\citeauthoryear{Premsri and Kordjamshidi}{2025}]{premsri2024neuro}
Tanawan Premsri and Parisa Kordjamshidi.
\newblock {Neuro-Symbolic Training for Reasoning over Spatial Language}.
\newblock In {\em NAACL}, 2025.

\bibitem[\protect\citeauthoryear{Qi \bgroup \em et al.\egroup }{2025}]{qi2025large}
Chengwen Qi, Ren Ma, Bowen Li, He~Du, Binyuan Hui, Jinwang Wu, Yuanjun Laili, and Conghui He.
\newblock Large language models meet symbolic provers for logical reasoning evaluation.
\newblock In {\em ICLR}, 2025.

\bibitem[\protect\citeauthoryear{Shao \bgroup \em et al.\egroup }{2024}]{shao2024chinatravel}
Jie-Jing Shao, Xiao-Wen Yang, Bo-Wen Zhang, Baizhi Chen, Wen-Da Wei, Lan-Zhe Guo, and Yu-feng Li.
\newblock China{T}ravel: A real-world benchmark for language agents in chinese travel planning.
\newblock {\em preprint arXiv:2412.13682}, 2024.

\bibitem[\protect\citeauthoryear{Shou \bgroup \em et al.\egroup }{2022}]{AMR-DA}
Ziyi Shou, Yuxin Jiang, and Fangzhen Lin.
\newblock {AMR-DA}: Data augmentation by abstract meaning representation.
\newblock In {\em ACL Findings}, pages 3082--3098, 2022.

\bibitem[\protect\citeauthoryear{Silver \bgroup \em et al.\egroup }{2016}]{alphago}
David Silver, Aja Huang, Chris~J Maddison, Arthur Guez, et~al.
\newblock Mastering the game of {Go} with deep neural networks and tree search.
\newblock {\em Nature}, 529(7587):484--489, 2016.

\bibitem[\protect\citeauthoryear{Su \bgroup \em et al.\egroup }{2025}]{dualformer}
DiJia Su, Sainbayar Sukhbaatar, Michael Rabbat, Yuandong Tian, and Qinqing Zheng.
\newblock Dualformer: Controllable fast and slow thinking by learning with randomized reasoning traces.
\newblock In {\em ICLR}, 2025.

\bibitem[\protect\citeauthoryear{Sur{\'\i}s \bgroup \em et al.\egroup }{2023}]{vipergpt}
D{\'\i}dac Sur{\'\i}s, Sachit Menon, and Carl Vondrick.
\newblock Viper{GPT}: Visual inference via {Python} execution for reasoning.
\newblock In {\em CVPR}, pages 11888--11898, 2023.

\bibitem[\protect\citeauthoryear{Taddeo and Floridi}{2005}]{TaddeoF05}
Mariarosaria Taddeo and Luciano Floridi.
\newblock Solving the symbol grounding problem: a critical review of fifteen years of research.
\newblock {\em JETAI}, 17(4):419--445, 2005.

\bibitem[\protect\citeauthoryear{Trinh \bgroup \em et al.\egroup }{2024}]{alphageometry}
Trieu~H Trinh, Yuhuai Wu, Quoc~V Le, He~He, and Thang Luong.
\newblock {Solving Olympiad Geometry without Human Demonstrations}.
\newblock {\em Nature}, 625(7995):476--482, 2024.

\bibitem[\protect\citeauthoryear{{\"U}nsal \bgroup \em et al.\egroup }{2024}]{alphaintegrator}
Mert {\"U}nsal, Timon Gehr, and Martin Vechev.
\newblock Alphaintegrator: Transformer action search for symbolic integration proofs.
\newblock {\em preprint arXiv:2410.02666}, 2024.

\bibitem[\protect\citeauthoryear{Weir \bgroup \em et al.\egroup }{2024}]{weir2024learning}
Nathaniel Weir, Muhammad Khalifa, Linlu Qiu, Orion Weller, and Peter Clark.
\newblock Learning to reason via program generation, emulation, and search.
\newblock In {\em NeurIPS}, 2024.

\bibitem[\protect\citeauthoryear{Weng \bgroup \em et al.\egroup }{2024}]{AutoCoNN}
Yixuan Weng, Minjun Zhu, Fei Xia, Bin Li, Shizhu He, Kang Liu, and Jun Zhao.
\newblock Mastering symbolic operations: Augmenting language models with compiled neural networks.
\newblock In {\em ICLR}, 2024.

\bibitem[\protect\citeauthoryear{Xu \bgroup \em et al.\egroup }{2018}]{SL}
Jingyi Xu, Zilu Zhang, Tal Friedman, Yitao Liang, and Guy Broeck.
\newblock A semantic loss function for deep learning with symbolic knowledge.
\newblock In {\em ICML}, 2018.

\bibitem[\protect\citeauthoryear{Yang \bgroup \em et al.\egroup }{2022}]{yang2022chain}
Mengjiao~Sherry Yang, Dale Schuurmans, Pieter Abbeel, and Ofir Nachum.
\newblock Chain of thought imitation with procedure cloning.
\newblock In {\em NeurIPS}, 2022.

\bibitem[\protect\citeauthoryear{Zhang \bgroup \em et al.\egroup }{2024}]{zhang2024dila}
Yu~Zhang, Hui-Ling Zhen, Zehua Pei, Yingzhao Lian, Lihao Yin, Mingxuan Yuan, and Bei Yu.
\newblock Dila: Enhancing {LLM} tool learning with differential logic layer.
\newblock {\em preprint arXiv:2402.11903}, 2024.

\bibitem[\protect\citeauthoryear{Zhou \bgroup \em et al.\egroup }{2025}]{zhou2025lawgpt}
Zhi Zhou, Kun-Yang Yu, Shi-Yu Tian, Jiang-Xin Shi, Xiao-Wen Yang, Pengxiao Song, Yi-Xuan Jin, Lan-Zhe Guo, and Yu-Feng Li.
\newblock Law{GPT}: Knowledge-guided data generation and its application to legal {LLM}.
\newblock {\em preprint arXiv:2502.06572}, 2025.

\bibitem[\protect\citeauthoryear{Zhou}{2019}]{abl}
Zhi{-}Hua Zhou.
\newblock Abductive learning: {T}owards bridging machine learning and logical reasoning.
\newblock {\em SCIS}, 62(7):76101, 2019.

\bibitem[\protect\citeauthoryear{Zhu \bgroup \em et al.\egroup }{2024}]{zhudeductive}
Tinghui Zhu, Kai Zhang, Jian Xie, and Yu~Su.
\newblock Deductive beam search: Decoding deducible rationale for chain-of-thought reasoning.
\newblock In {\em COLM}, 2024.

\end{thebibliography}
\end{document}